\documentclass{article}
\usepackage{ijcai20}

\usepackage{microtype}
\usepackage[utf8]{inputenc}
\usepackage{comment}
\usepackage{times}
\usepackage[utf8]{inputenc}
\usepackage[small]{caption}
\usepackage{url}
\usepackage[hidelinks]{hyperref}
\usepackage{soul}

\usepackage{graphicx, lipsum, wrapfig}
\usepackage{amsmath,amssymb,amsthm}
\usepackage{mathtools}
\usepackage{IEEEtrantools}
\theoremstyle{plain}

\theoremstyle{definition}

\providecommand\given{}
\newcommand\SetSymbol[1][]{\nonscript\:#1\vert\nonscript\:
  \mathopen{}\allowbreak}
\DeclarePairedDelimiterX\Set[1]\{\}{%
  \renewcommand\given{\SetSymbol[\delimsize]}
  #1 }
\DeclarePairedDelimiterX\paren[1](){%
  \renewcommand\given{\SetSymbol[\delimsize]}
  #1 }

\DeclareMathOperator*{\E}{\mathbb{E}}

\usepackage[ruled,vlined,linesnumbered]{algorithm2e}
\newcommand{\algorithmfootnote}[2][\footnotesize]{%
  \let\old@algocf@finish\@algocf@finish
  \def\@algocf@finish{\old@algocf@finish
    \leavevmode\rlap{\begin{minipage}{\linewidth}
    #1#2
    \end{minipage}}%
  }%
}

\newcommand{\blue}[1]{\textcolor{black}{#1}}

\newcommand{\cyan}[1]{\textcolor{black}{#1}}

\usepackage{todonotes}%
\newcommand{\jae}[1]{%
  \todo[%
  author=Jae,%
  inline,%
  caption={},%
  size=footnotesize,%
  color=yellow!10!white,%
  textcolor=black,%
  bordercolor=black%
  ]{#1}%
}

\usepackage{pgfplots}
\usepgfplotslibrary{groupplots}
\usepackage{xspace}
\usepackage{adjustbox}
\usepackage{textcomp}
\usepackage{mathtools}
\usepackage{multirow}
\usepackage{cutwin}

\usepackage{tikz}
\usetikzlibrary{automata,arrows,shapes}
\usepackage{mathrsfs}
\pgfdeclarelayer{background}
\pgfsetlayers{main,background}
\tikzstyle{vertex}=[circle,fill=black!10,minimum size=20pt,inner sep=0pt]
\tikzstyle{edge} = [draw,thick]
\tikzstyle{weight} = [font=\small]
\tikzstyle{redundant edge} = [draw,line width=5pt,-,red,opacity=.7]

\pgfplotsset{width=.30\linewidth,compat=1.9}

\title{Deep Hurdle Networks for Zero-Inflated Multi-Target Regression: \\ Application to Multiple Species Abundance Estimation}

\author{
Shufeng Kong$^1$\footnote{Contact Author}
\and
Junwen Bai$^1$\and
Jae Hee Lee$^2$ \and
Di Chen$^1$ \and
Andrew Allyn$^3$ \and \\
Michelle Stuart$^4$ \and
Malin Pinsky$^4$ \and
Katherine Mills$^3$ \and
Carla P. Gomes$^1$
\affiliations
$^1$Department of Computer Science, Cornell University, USA \\
$^2$School of Computer Science and Informatics, Cardiff University, UK \\
$^3$Gulf of Maine Research Institute, USA\\
$^4$Department of Ecology, Evolution, and Natural Resources, Rutgers University, USA 
\emails 
\{sk2299, jb2467, dc874\}@cornell.edu,
gomes@cs.cornell.edu,
leejh3@cardiff.ac.uk, \\
\{aallyn, kmills\}@gmri.org,
\{michelle.stuart, malin.pinsky\}@rutgers.edu
}

\begin{document}

\maketitle

\begin{abstract}
  A key problem in computational sustainability is to understand the
  distribution of species across landscapes over time. This question
  gives rise to challenging large-scale prediction problems since (i)
  hundreds of species have to be simultaneously modeled and (ii) the
  survey data are usually inflated with zeros due to the absence of
  species for a large number of sites. The problem of tackling
  \textit{both} issues simultaneously, which we refer to as the
  \emph{zero-inflated multi-target regression problem}, has not been
  addressed by previous methods in statistics and machine learning.
  In this paper, we propose a novel deep model for the zero-inflated
  multi-target regression problem. To this end, we first model the
  joint distribution of multiple response variables as a multivariate
  probit model and then couple the positive outcomes with a
  multivariate log-normal distribution. 
  \blue{By penalizing the difference between the two distributions' covariance matrices,}
  a link between both
  distributions is established. The whole model is cast as an end-to-end
  learning framework and we provide an efficient learning algorithm
  for our model that can be fully implemented on GPUs. We show that
  our model outperforms the existing state-of-the-art baselines on two
  challenging real-world species distribution datasets concerning bird
  and fish populations.
\end{abstract}

\section{Introduction}

Since the Industrial Revolution there has been an increase in
biodiversity loss, due to a combination of factors such as
agriculture, urbanization, and deforestation, as well as climate
change and human introduction of non-native species to ecosystems.
Biodiversity loss is a great challenge for humanity, given the
importance of biodiversity for sustaining ecosystem services. For
example, bird species play a key role in regulating ecosystems by
controlling pests, pollinating flowers, spreading seeds and
regenerating forests. In a current study it was shown that the bird population in the United States and Canada has fallen by an estimated $29\%$
since 1970~\cite{rosenberg2019decline}. The biomass of top marine
predators \blue{has declined substantially in many cases,
\cite{mccauley2015marine}, 
and many marine species are shifting rapidly to new regions in response to changing ocean conditions \cite{pinsky2020climate}}. More generally, a recent report from the
United Nations warned that about a million animal and plant species
face extinction 
~\cite{brondizio2019global}.

To protect and conserve species, a key question in computational
sustainability concerns understanding the distribution of species
across landscapes over time, which gives rise to challenging
large-scale spatial and temporal modeling and prediction
problems~\cite{gomes2019computational}. In particular, ecologists are
interested in understanding how species interact with the environment
as well as how species interact with each other. 
Joint species distribution modeling is therefore a computational challenge, as we are interested in simultaneously modeling the correlated distributions of potentially hundreds of species, rather than a single species at a time as traditionally done.
Another challenge in joint species distribution modeling is that often outcomes of interest, such as local species abundance in terms of counts or biomass, are sparsely observed, leading to zero-inflated data \cite{10.1371/journal.pone.0196127}.
Zero-inflated data are frequent in many other
settings beyond ecology, for example in research studies concerning
public health when counting the number of vaccine adverse events
\cite{rose2006use} and prediction of sparse user-item consumption
rates \cite{lichman2018prediction}.

Herein we propose general models for jointly estimating counts or
abundance for multiple entities, which are referred to as
\textit{zero-inflated multi-target regression}. While
our models are general, we focus on computational sustainability
applications, in particular the joint estimation of counts of multiple
bird species and biomass of multiple fish species. As discussed above,
\blue{many of} these species have suffered dramatic reductions  \blue{or changes in geographic distributions} in recent years.

Our contributions are:
\begin{enumerate}
\item We propose a deep generalization of the conventional model for
  multi-target regression that simultaneously models zero-inflated
  data and the correlations among the multiple response
  variables. 
\item We provide an efficient end-to-end learning framework for our
  model that can be implemented on GPUs.
\item We evaluate our model as well as state-of-the art joint
  distribution models on two datasets from computational
  sustainability that concern the distributions of fish and bird
  species. 
\end{enumerate}


\section{Related Work}\label{sec:other_works}

A popular model for zero-inflated data is the so-called
\emph{hurdle model} \cite{mullahy1986specification} in which a
Bernoulli distribution governs the binary outcome of whether a
response variable has a zero or positive realization; if the
realization is positive, the ``hurdle'' is crossed, and the
conditional distribution of the positives is governed by a
zero-truncated parametric distribution. Although the hurdle model is
popular for zero-inflated data, it has \blue{several} limitations: 1) \blue{the} two
components of the model are assumed to be independent; 2) \blue{it} does not
explicitly capture the relationships between multiple response variables; \blue{and}
3) \blue{its} traditional implementation does not scale. In our
work, we address these three key limitations. 

A closely related alternative to handle zero-inflated count data is
the family of \emph{zero-inflated models}
\cite{lichman2018prediction}, \blue{in which}
the response variable is also modeled as a mixture of a Bernoulli
distribution and a parametric distribution on non-negative integers
such as the Poisson or negative binomial distributions. Different from
hurdle model, the \blue{conditional} distribution of a
zero-inflated model is not required to be zero-truncated. In other
words, while the hurdle model assumes zeros only come from the
Bernoulli distribution, a zero-inflated model assumes zeros could come
from both the Bernoulli distribution and the \blue{conditional} distribution.
To account for the inherent correlation of response variables, a class
of \emph{multi-level zero-inflated regression models} was presented
\cite{almasi2016multilevel}. With such a model, response variables are
organized in a hierarchical structure. Response variables are taken to
be independent between clusters. Cluster level and within-cluster
correlations of response variables are modeled explicitly through
random effects attached to linear predictors. However, characterising
a suitable hierarchical structure is nontrivial\blue{, and} these models \blue{also} do
not explicitly capture \blue{covariance relations} between response variables.

The general problem of multiple-target regression has been extensively
studied
\cite{borchani2015survey,xi2018empirical,DBLP:journals/corr/abs-1901-00248}.
Existing methods for multi-output regression can be categorized as:
(1) \blue{p}roblem transformation methods that transform the multi-output
problem into independent single-output problems each solved using a
single-output regression algorithm, such as multi-target regressor
stacking (MTRS) \cite{spyromitros2016multi} \blue{and} regressor chains (RC)
\cite{melki2017multi}\blue{; and (2)} algorithm adaptation methods that adapt a
specific single-output method to directly handle multi-output dataset,
such as multi-objective random forest (MORF)
\cite{kocev2007ensembles}, random linear target combination (RLTC)
\cite{tsoumakas2014multi}\blue{, and} multi-output support vector regression
(MOSVR) \cite{zhu2018efficient}. An empirical comparison \blue{of} three
representative state-of-the-art multi-output regression learning
algorithms, MTRS, RLTC and MORF, is presented in
\cite{xi2018empirical}. Although these advanced multiple-output
regression algorithms exploit some correlation between response
variables to improve \blue{predictive} performance, they do not fully model the
covariance structure of the response variables and do not consider
zero-inflated data. Our work explicitly models zero-inflated data in
multi-target regression and it also explicitly models the covariance
underlying the phenomena to better characterize the correlation among
entities.


\section{Preliminaries}

In this section we give a brief introduction to the hurdle model and
multivariate probit model. In the following we use $\mathbb{R}$,
$\mathbb{R}_0$, $\mathbb{R}_+$, $\mathbb{N}$, $\mathbb{N}_0$ and
$\mathbb{N}_+$ to denote the reals, nonnegative reals, positive reals,
integers, nonnegative integers and positive integers, respectively. If
\blue{an} $L$-dimensional vector $y$ is integral, we write
$y\in \mathbb{N}^{L}$. We write $y\in \mathbb{R}^{L}$ if $y$ is real.
Two vectors $x,y\in \mathbb{R}^{L}$ satisfy $x \preccurlyeq y$ (or
$x \succcurlyeq y$) iff $x_j \le y_j$ (or $x_j \ge y_j$) for
$1\le j \le L$. Since we consider the problem of species abundance
estimation, in the following all label data are assumed to be
nonnegative. We denote the probability density function (PDF) and the
cumulative density function (CDF) of a
multivariate normal distribution $\mathcal{N}(\mu,\Sigma)$ as
$\phi\paren{x \given \mu,\Sigma}$ and $\Phi\paren{x \given \mu, \Sigma}$, respectively.

\subsection{Hurdle Model}

The hurdle model aims to fit data with two independent distributions:
(1) \blue{a} Bernoulli distribution which governs the binary outcome of a
response variable being zero or positive; \blue{and} (2) a zero-truncated
distribution of the \blue{positive} response variable. Specifically, given a dataset
$D = \Set{ (x, y)^{(i)} \given i=1,\ldots, N }$, where
$x \in \mathbb{R}^M$ is the feature data and $y\in\mathbb{R}^L_{0}$
(or $y \in \mathbb{N}_0^L$) is the label data, and let $p_{j}$ be the
probability of $y_{j}$ $(1\le j \le L)$ being positive, then we have
$y_{j}' \sim \text{Bernoulli}(p_{j})$, where $y_{j}' = 1$ if $y_{j}>0$
and $0$ otherwise. Let $f\paren{y_{j} \given y_{j}>0}$ be the PDF of
the zero-truncated distribution of $y_j$. The likelihood of $y_j$
is given as
$\mathcal{L}(y_{j}) = \Pr(y_{j}'\blue{=1})f\paren{y_{j} \given y_{j}>0}$.

\subsection{Multivariate Probit Model}

Given a dataset $D = \Set{ (x, y')^{(i)} \given i=1,\ldots, N }$, where
$x \in \mathbb{R}^M$ is the feature data and $y'\in\{0,1\}^L$ is the
\blue{present/absent} label data, the \emph{multivariate probit model} (MVP)
\cite{chen2018end} maps the Bernoulli distribution of each binary
response variable $y_{j}'$ to a latent variable $r_{j} \in \mathbb{R}$
through threshold $0$, where $r = (r_{1},\ldots, r_{L})$ is subject to
a multivariate normal distribution:
\begin{align}
    \Pr\paren{y_{j}'=1 \given x} &=  \Pr\paren{r_{j} > 0 \given x} \label{eq:mvp_map}\\
    \Pr\paren{y_{j}'=0 \given x} &= \Pr\paren{r_{j} \le 0 \given x}\label{eq:mvp_map2}
\end{align}
where $r \sim \mathcal{N}(\mu, \Sigma)$. The joint likelihood of $y'$
is given as
\begin{equation}\label{mul_normal_cdf}
  \mathcal{L}\paren{y' \given x} = \int_{A_1}\cdots \int_{A_L} \phi\paren{r \given \mu,\Sigma}\text{ d}r_{1},\cdots, \text{d}r_{L}
\end{equation}
where $A_j=(-\infty,0]$ if $y_{j}'=0$, $A_j=[0, \infty)$ if
$y_{j}'=1$. Although there is no closed-form expression for the CDF of a
general multivariate normal distribution, \cite{chen2018end} proposed
an efficient parallel sampling algorithm to approximate it. We can
first translate equation (\ref{mul_normal_cdf}) into the CDF of a
multivariate normal distribution using the affine transformation:
\begin{equation}\label{mul_normal_affine}
  \mathcal{L}\paren{y' \given x} = \Phi\paren{0 \given -\mu', \Sigma'}
\end{equation}
where $\mu'=U\mu$, $\Sigma'=U\Sigma U$, and
$U=\mbox{diag}(2y'-1)\in \{-1, 0, 1\}^{L \times L}$ is the diagonal
matrix with vector $2y'-1$ as its diagonal. By decomposing $\Sigma'$
into $I + \Sigma''$, where $I$ is the identity matrix, a random
variable $r \sim \mathcal{N}(0, \Sigma')$ can be written as $r=z-w$,
where $z\sim \mathcal{N}(0, I)$ and $w\sim \mathcal{N}(0,\Sigma'')$.
Then, $\Phi\paren{0 \given -\mu', \Sigma'}$ can be approximated as

\begin{IEEEeqnarray}{lCl}
  \Phi\paren{0 \given -\mu', \Sigma'} &=& \Pr(r-\mu' \preccurlyeq 0) =
  \Pr(z-w \preccurlyeq \mu') \nonumber\\
  \IEEEeqnarraymulticol{3}{l}{= \E_{w\sim
    \mathcal{N}(0, \Sigma'')} [\Pr\paren{z \preccurlyeq
    (w+\mu')\given w}]} \nonumber\label{cdf_approx}\\
  \IEEEeqnarraymulticol{3}{l}{= \E_{w\sim \mathcal{N}(\mu', \Sigma'')} 
  [\prod_{j=1}^L
    \Phi(w_j)]}\\
  \IEEEeqnarraymulticol{3}{l}{
  \approx \frac{1}{K}\sum_{k=1}^K \prod_{j=1}^L
  \Phi(w_j^{(k)})}\nonumber
\end{IEEEeqnarray}
where samples $\{w^{(k)}\}$ are subject to
$\mathcal{N}(\mu',\Sigma'')$ and $\Phi$ is the CDF of the standard
normal distribution. Note that according to (\ref{mul_normal_affine}),
we have $\Sigma''=\Sigma'-I=U\Sigma U-I =
U(\Sigma-I)U$. 
 \begin{figure}[t]
   \centering \includegraphics[width=0.99\linewidth]{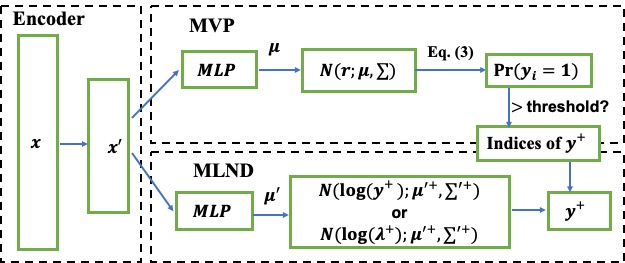}
   \caption{The deep hurdle network architecture. An encoder is used
     to learn latent features $x'$, an MVP is used to model
     the joint distribution of multiple response variables $y$ being
     zero or positive, and an MLND is used to model the joint
     distribution of positive response variables $y^+$. The MVP and
     MLND are linked by sharing the same latent features $x'$, \blue{and penalizing the difference between their covariance matrices $\Sigma$ and $\Sigma'$}. 
   }\label{fig:hurdle_net}
 \end{figure}

\section{Deep Hurdle Network}

\blue{In this paper we provide a deep generalization of the hurdle model, within an autoencoder framework, which we call the deep hurdle network (DHN). The DHN integrates the MVP to model the joint distribution of multiple response variables being zero or positive, and the multivariate log-normal distribution (MLND) to model the positive response variables. The MVP and MLND share the same latent features, and differences between their covariance matrices are penalized.}
Specifically, given a dataset $D =  \Set{(x, y)^{(i)} \given
  i=1,\ldots, N} $, where $x \in \mathbb{R}^M$ is the feature data and
$y\in\mathbb{R}^L_{0}$ (or $y \in\mathbb{N}_0^L$) is the label data, the hurdle network contains three parts (\blue{see}~Figure~\ref{fig:hurdle_net} for an illustration):
\begin{itemize}
\item[1.] \textbf{Encoder}: An encoder maps the raw features
$x\in \mathbb{R}^M$ to latent features $x'\in \mathbb{R}^{\blue{M'}}$.

\item[2.] \textbf{MVP}: Label data $y\in \mathbb{R}^L_0/\mathbb{N}^L_0$ \blue{are}
translated into binary label data $y'\in \{0,1\}^L$, where
$y'=(y_{1}', \ldots, y_{L}')$ and $y_{j}'$ equals to $1$ if $y_{j}>0$
and $0$ otherwise. We then use MVP to map $y'$ to latent variable
$r\in \mathbb{R}^L$ via equations \eqref{eq:mvp_map}--\eqref{eq:mvp_map2}, where $r$ is
assumed to follow $\mathcal{N}(\mu,\Sigma)$. A multilayer perceptron
(MLP) is then used to model $\mu$ given $x'$ as input. $\Sigma$ is a
global parameter which is learned from random initialization and shared
by all data points.

\item[3.] \textbf{MLND}: \blue{Let
$y^+ \in \mathbb{R}_+^P$ $(P\le L)$ be the positive
part of $y\in \mathbb{R}^L$ \blue{, where $P$ is the number of positive elements of $y$}.
$\log(y^+)=(\log(y_{1}^+),\ldots,\log(y_{P}^+))$ is directly modeled
as a multivariate normal distribution, \blue{and} $y^+$ are assumed to
follow a multivariate log-normal distribution. Therefore, we
have}
\begin{equation}\label{MDM}
  \begin{split}
    \mathcal{L}\paren{\log(y^+) \given x'} = \phi\paren{\log(y^+)\given{\mu'}^+,{\blue{\Sigma'}}^+}
  \end{split}
\end{equation}

\blue{where ${\mu'}^+$ and
${\Sigma'}^+$ are the sub-parts of $\mu'\in \mathbb{R}^L$ and
$\Sigma'$ that correspond to
$y^+$ respectively. Note that here
$\Sigma'$ is} 
\blue{encouraged to be similar to the convariance matrix} 
\blue{$\Sigma$ of the MVP and $\mu' \in
\mathbb{R}^L$ is modeled with an MLP.}\\

\blue{On the other hand, let
$y^+ \in \mathbb{N}^P_+(P\le L)$ be the positive part of $y\in \mathbb{N}^L$. Each $y_{j}^+$ is assumed to follow an
univariate Poisson distribution
$\Pr\paren{y^+_{j} \given \lambda_{j}^+}$ for $1\le j \le P$, where
$\lambda_{j}^+$ is the mean, and
$\log(\lambda^+)=(\log(\lambda_{1}^+),\ldots,\log(\lambda_{P}^+))$ is
assumed to follow a multivariate normal distribution. Therefore, we
have}
\begin{equation}\label{MDM1}
  \begin{split}
    \mathcal{L}\paren{\log({\lambda}^+) \given x'} = \phi\paren{\log({\lambda}^+)\given{\mu'}^+,{\blue{\Sigma'}}^+}
  \end{split}
\end{equation}

\end{itemize}

There are several advantages of the deep hurdle network over the
conventional hurdle model:
\begin{itemize}

\item[1.] The encoder is forced to learn the salient features and ignore the
noise and irrelevant parts of the raw features. This relieves us from
selecting \blue{which} salient features to use in the conventional hurdle
model.%

\item[2.] DHN adopts MVP and MLND to handle correlations between multiple
response variables explicitly via \blue{covariance matrices}, which is not
considered in the conventional hurdle model.

\item[3.] The two components of the conventional hurdle model are
independent, while in DHN the MVP and MLND are linked by sharing the
same latent features, \blue{and penalizing the different between their covariance matrices.}%
\end{itemize}
%

\subsection{End-to-End Learning for DHN}

Parameters of a deep model are usually estimated by minimizing the
negative log-likelihood (NLL). We develop two different
learning objectives for nonnegative continuous and count data,
respectively.
After selecting our objective function, we can estimate the parameters of the deep hurdle network by minimizing the objective function via stochastic gradient descent (SGD).
\subsubsection{Learning Objective for Nonnegative Continuous Data}
If response variables $y$ are nonnegative reals, then
we combine equations (\ref{mul_normal_affine}), (\ref{cdf_approx}) and
(\ref{MDM}) to obtain the negative log-likelihood (NLL) function:
\begin{IEEEeqnarray}{l}
  -\log(\mathcal{L}\paren{y'\given x'}\mathcal{L}\paren{\log(y^+) \given x'})\nonumber\\
= -\log\left(\E_{w \sim \mathcal{N}(U\mu,U(\Sigma-I) U)}\left[\prod_{j=1}^L \Phi(w_j)\right]\right) \label{obj_continous_tmp}\\
\IEEEeqnarraymulticol{1}{l}{\phantom{=}-\log(\phi\paren{\log(y^+) \given {\mu'}^+,{\Sigma'}^+})} \nonumber \label{real_obj_real}
\end{IEEEeqnarray}
The first part of \cyan{the right-hand side of} equation (\ref{obj_continous_tmp}) can be approximated by a
set of samples $\{w^{(k)}\}$ from $\mathcal{N}(U\mu,U(\Sigma-I)U)$:
$-\log(\frac{1}{K}\sum_{k=1}^K \exp(\sum_{j=1}^L \log(\Phi(w_j^{(k)}))))$.
Such a set of samples can be obtained by first using the
\emph{Cholesky decomposition}, which decomposes $\Sigma-I$ into
$CC^T$, and then generating independent samples
$v^{(k)}\sim \mathcal{N}(0,I)$ \blue{to yield} $w^{(k)}=U(\mu + Cv^{(k)})$.
According to the affine transformation of the normal distribution, the
samples $\{w^{(k)}\}$ are subject to $\mathcal{N}(U\mu,U(\Sigma-I)U)$.
However, doing the matrix multiplications $U\mu$ and $UCv^{(k)}$ are
unnecessary since $U=\mbox{diag}(2y'-1)\in \{-1, 0, 1\}^{L \times L}$
is the diagonal matrix with vector $2y'-1$ as its diagonal. Let
$w'^{(k)}=\mu + Cv^{(k)}$ such that $\{w'^{(k)}\}$ are subject to
$\mathcal{N}(\mu,\Sigma-I)$, then it is easy to show that
$\log(\Phi(w_j^{(k)})) = y_{j}'\log(\Phi(w'^{(k)}_j)) +
(1-y_{j}')\log(1-\Phi(w'^{(k)}_j))$.
\blue{Thus, we can approximate equation (\ref{obj_continous_tmp}) as follows:} 
\begin{IEEEeqnarray}{l}\label{final_real_obj}
    - \log\Bigl(\frac{1}{K}\sum_{k=1}^K\exp\Bigl(\sum_{j=1}^L
    (y_{j}'\log(\Phi(w^{(k)}_j))\label{real_nll_final}\\
    \IEEEeqnarraymulticol{1}{r}{{}+ (1-y_{j}')\log(1-\Phi(w^{(k)}_j)))\Bigl)\Bigl)}\nonumber\\
    {-\log(\phi\paren{\log(y^+) \given {\mu'}^+,{\Sigma'}^+})}\nonumber
\end{IEEEeqnarray}
\blue{where samples $\{w^{(k)}\}$ are subject to
$\mathcal{N}(\mu,\Sigma-I)$.}
\subsubsection{Learning Objective for Count Data}
If response variables $y$ are nonnegative integers, let
$y^+ \in \mathbb{N}_+^P$ be the positive part of
$y\in \mathbb{N}_0^L(P \le L)$, \blue{where $P$ is the number of positive elements of $y$}, then each $y_{j}^+$ is assumed to
follow a univariate Poisson distribution $\Pr(y_{j}^+;\lambda_{j}^+)$,
and $\log(\lambda^+)=(\log(\lambda_{1}^+),\ldots,\log(\lambda_{P}^+))$
is assumed to follow a multivariate normal distribution
$\mathcal{N}({\mu'}^+, {\Sigma'}^+)$. The likelihood of response
variables $y^+$ is given as
\begin{equation}\label{postive_response_var_llh}
  \begin{split}
    &\mathcal{L}(y^+;\lambda^+|\mathcal{N}(\log(\lambda^+);\mu',{\Sigma'}^+), x')\\
    &= \E_{\log(\lambda^+)\sim \mathcal{N}({\mu'}^+,{\Sigma'}^+)}
    [\prod_{j=1}^P
    \frac{{\lambda_{j}^+}^{y_{j}^+}e^{-\lambda_{j}^+}}{y_{j}^+!}]
  \end{split}
\end{equation}

We then combine equations (\ref{mul_normal_affine}),
(\ref{cdf_approx}) and (\ref{postive_response_var_llh}) to obtain the
NLL function as follows:

\begin{IEEEeqnarray}{l}\label{final_inter_obj}
  -\log(\mathcal{L}(y'|x')\mathcal{L}(y^+;\lambda^+|\mathcal{N}(\log(\lambda^+);\mu',{\Sigma'}^+), x')\nonumber\\
  = -\log(\E_{w \sim \mathcal{N}(L\mu,L(\Sigma-I) L)}[\prod_{j=1}^L \Phi(w_j)]) \nonumber\\
  \IEEEeqnarraymulticol{1}{r}{- \log(\E_{\log(\lambda^+)\sim \mathcal{N}({\mu'}^+,{\Sigma'}^+)} [\prod_{j=1}^P \frac{{\lambda_{j}^+}^{y_{j}^+}e^{-\lambda_{j}^+}}{y_{j}^+!}])}\nonumber\\
  \approx - \log(\frac{1}{K}\sum_{k=1}^K(\exp(\sum_{j=1}^L  (y_{j}'\log(\Phi(w^{(k)}_j))) \nonumber\\
  \IEEEeqnarraymulticol{1}{r}{+ (1-y_{j}')\log(1-\Phi(w^{(k)}_j)))))}\nonumber\\
  -\log(\frac{1}{K}\sum_{k=1}^K\exp(\sum_{j=1}^P
  (y_{j}^+\log({\lambda^+})_j^{(k)} -
  \lambda_{j}^{(k)}-\log(y_{j}^+!)))) \nonumber\\
  \label{integer_nll_final}
\end{IEEEeqnarray}
where samples $\{w^{(k)}\}$ and $\{\log(\lambda^+)^{(k)}\}$ are
subject to $\mathcal{N}(\mu,\Sigma-I)$ and
$\mathcal{N}({\mu'}^+,{\Sigma'}^+)$, respectively.

\blue{Lastly, we add an L1 loss term to  equations (\ref{final_real_obj}) and (\ref{final_inter_obj}), respectively, to penalize the difference between $\Sigma$ and $\Sigma'$.}

\section{Experiments}
In this section, we compare our model with the existing state-of-the-art baselines on two challenging real-world species distribution datasets concerning bird and fish populations.

\subsection{Datasets}

\begin{figure}
    \begin{minipage}{.49\linewidth}
      \centering
      \includegraphics[width=1.1\linewidth]{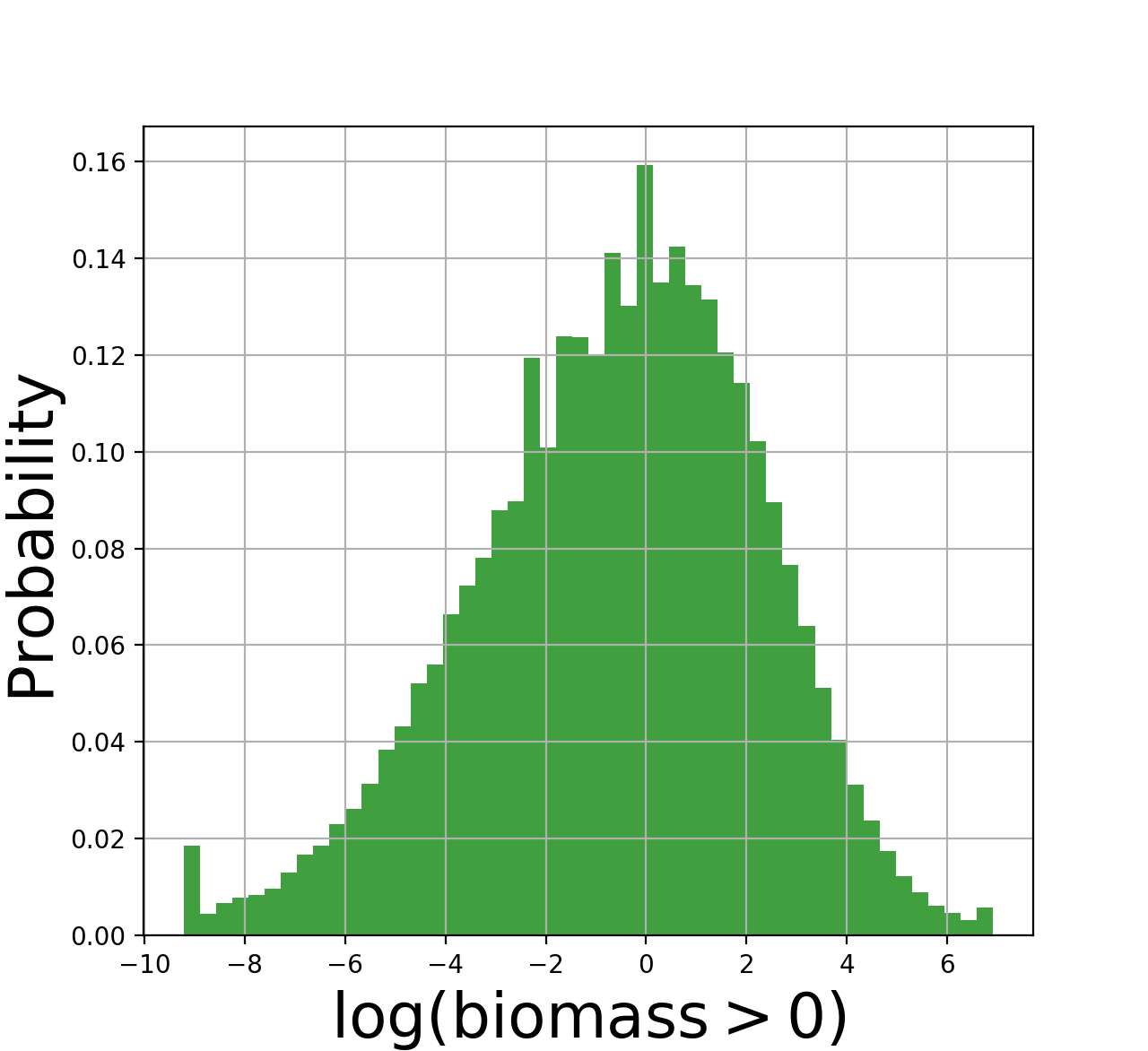}
    \end{minipage}
    \begin{minipage}{.49\linewidth}
      \centering
      \includegraphics[width=1.1\linewidth]{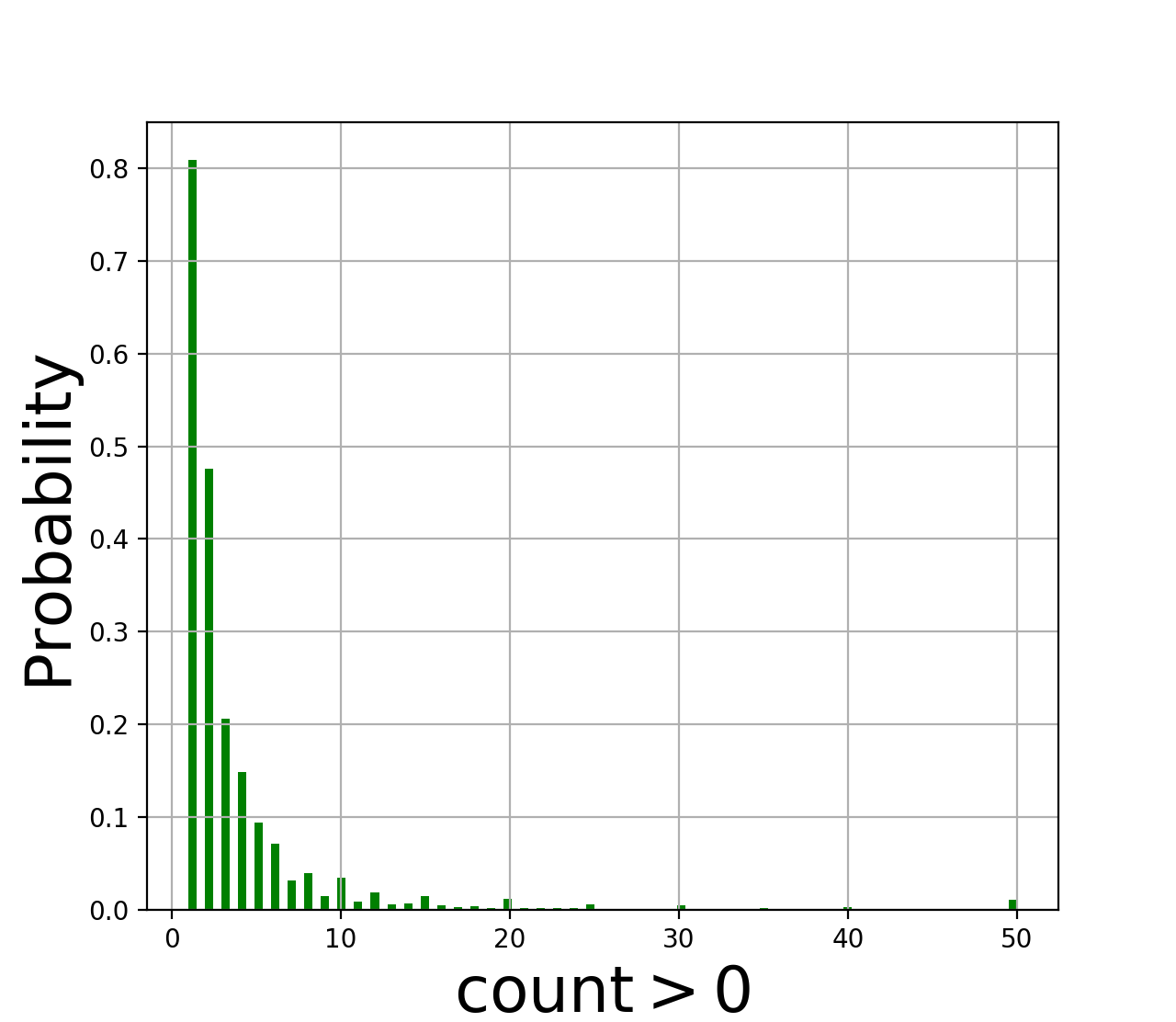}
    \end{minipage}
    \caption{Visualization of data distribution. \blue{The left figure shows that the sea bottom trawl survey data approximately follow a log-normal distribution, and the right figure shows that the eBird data approximately follow a Poission distribution.}    \label{fig:data_distribution}}
\end{figure}

\textbf{Sea bottom trawl surveys (SBTSs)} are scientific surveys that collect data on the distribution and abundance of marine fishes.
In each haul (sample), all catches of each species were weighted and recorded, as well as the time, location, sea surface temperature and depth.
For each haul, we have additional environmental features from the Simple Ocean Data Assimilation (SODA) Dataset \cite{carton2018soda3} and Global Bathymetry and Elevation Dataset 
, such as sea bottom temperature, rugosity, etc. Thus, we have a 17-dimensional feature vector for each haul.
We study fish species abundance in the continental shelf around North America and consider SBTSs
that were conducted from 1963 to 2015 
\cite{10.1371/journal.pone.0196127}. We consider the top 100 most frequently observed species as the target species and there are $135,458$ hauls that caught at least one of these species. \blue{Among} the $135,458*100$ data entries, only $8.5\%$ of them are nonzero. The distribution of the positive data  is visualized in Figure~\ref{fig:data_distribution}.




\textbf{eBird} is a crowd-sourced bird observation dataset \cite{munson2009ebird}. A record in this dataset corresponds to a checklist that an experienced bird observer uses to mark the number of birds of each species detected, as well as the time and location.
Additionally, we obtain a 16-dimensional feature vector for each observation location from the National Land Cover Dataset (NLCD) \cite{homer2015completion} which describes the landscape composition with respect to 16 different land types such as water, forest, etc.
We \blue{use} all the bird observation checklists in North America in the last two weeks of May from 2002 to 2014. We consider the top 100 most frequently observed species as the target species and there are $39,668$ observations that contain at least one of these 100 species. Only $19.8\%$ of these $39,668*100$ data entries are nonzero. The distribution of the positive data is visualized in Figure~\ref{fig:data_distribution}.


\subsection{Evaluation Metrics}

For the task of multi-target regression, we employ and adapt two well
known measures: the average correlation coefficient (ACC)\footnote{
  The values range between -1 and 1, where -1 shows a
  perfect negative correlation, while a 1 shows a
  perfect positive correlation.}%
and root mean squared error (RMSE) \cite{borchani2015survey,DBLP:journals/corr/abs-1901-00248}. The ACC is defined as:
$$\frac{1}{L}\sum_{j=1}^L \frac{\sum_{i=1}^{N_{\text{test}}}(y_j^{(i)}-\bar{y}_j)(\hat{y}_j^{(i)}-\bar{\hat{y}}_j)}{\sqrt{\sum_{i=1}^{N_{\text{test}}}(y_j^{(i)}-\bar{y}_j)^2\sum_{i=1}^{N_{\text{test}}}(\hat{y}_j^{(i)}-\hat{\bar{y}}_j)^2}},$$
 where $L$ is the number of response variables, $N_{\text{test}}$ is the number of testing samples, $y^{(i)}$ and $\hat{y}^{(i)}$ are the vectors of the actual and predicted outputs for $x^{(i)}$, respectively, and $\bar{y}$ and $\bar{\hat{y}}$ are the vectors of averages of the actual and predicted outputs, respectively.
Since the data considered here \blue{are}
zero-inflated, using standard RMSE \blue{might} not be appropriate. Models
can produce degenerate results by simply predicting a near-zero vector for
each test data point. Therefore, we adapt the RMSE to the zero-inflated
setting by considering zero and positive parts of \blue{the} output
separately. Let $y^{(i)}$ be the actual $L$-dimensional output vector
of the $i$-th testing point, $I_0$ (respectively, $I_+$) be the set of indices
of zero (respectively, positive) \blue{elements} in $y^{(i)}$, and $\hat{y}$ be the
predicted $L$-dimensional output vector, then we define the
zero-inflated RMSE (zRMSE) as:
$$
\frac{\sum_{i=0}^{N_{\text{test}}} \sqrt{\frac{\alpha\sum_{j\in I_0}(\hat{y}^{(i)}_{j})^2}{|I_0|} + \frac{(1-\alpha)\sum_{j\in I_+}(y^{(i)}_{j}-\hat{y}^{(i)}_{j})^2}{|I_+|}}}{N_{\text{test}}}, 
$$
where $0\le \alpha \le 1$ is the relative importance of the zero part. 

From the above new definition, we can see that the zero and positive parts
of an output are both considered, therefore, a model cannot cheat by
predicting near-zero vectors by ignoring the positive parts of
the outputs. Finally, we also compare the time to learn different models.

\subsection{Baselines}

We consider both the state-of-the-art hurdle/zero-inflated models in \blue{statistics} and multi-target regression models in machine learning:

1) Hurdle/zero-inflated models. 
For the bird counting data, we select the state-of-the-art multi-level zero-inflated Poisson model (MLZIP) \cite{almasi2016multilevel} as a baseline. For the fish biomass data, we modify the MLZIP by replacing the Poisson with log-normal, which we denote as multi-level zero-inflated log-normal (MLZILN) model, as a baseline. For both the baselines, response variables are divided into random clusters. 



2) Multi-target regression models. As we have discussed in Section~\ref{sec:other_works}, there are two kinds of methods for multi-target regression models: \blue{for} problem transformation methods, we select the state-of-the-art multi-target regressor stacking (MTRS) \cite{spyromitros2016multi} as a baseline; \blue{and,} for algorithm adaptation methods, we select the state-of-the-art multi-objective random forest (MORF) \cite{kocev2007ensembles}, random linear target
combination (RLTC) \cite{tsoumakas2014multi}, multi-output support vector regression (MOSVR) \cite{zhu2018efficient}, and multi-layer multi-target regression (MMR) \cite{zhen2017multi} as baselines.

\begin{figure}
    \begin{minipage}{.50\linewidth}
      \centering \includegraphics[width=1.09\linewidth]{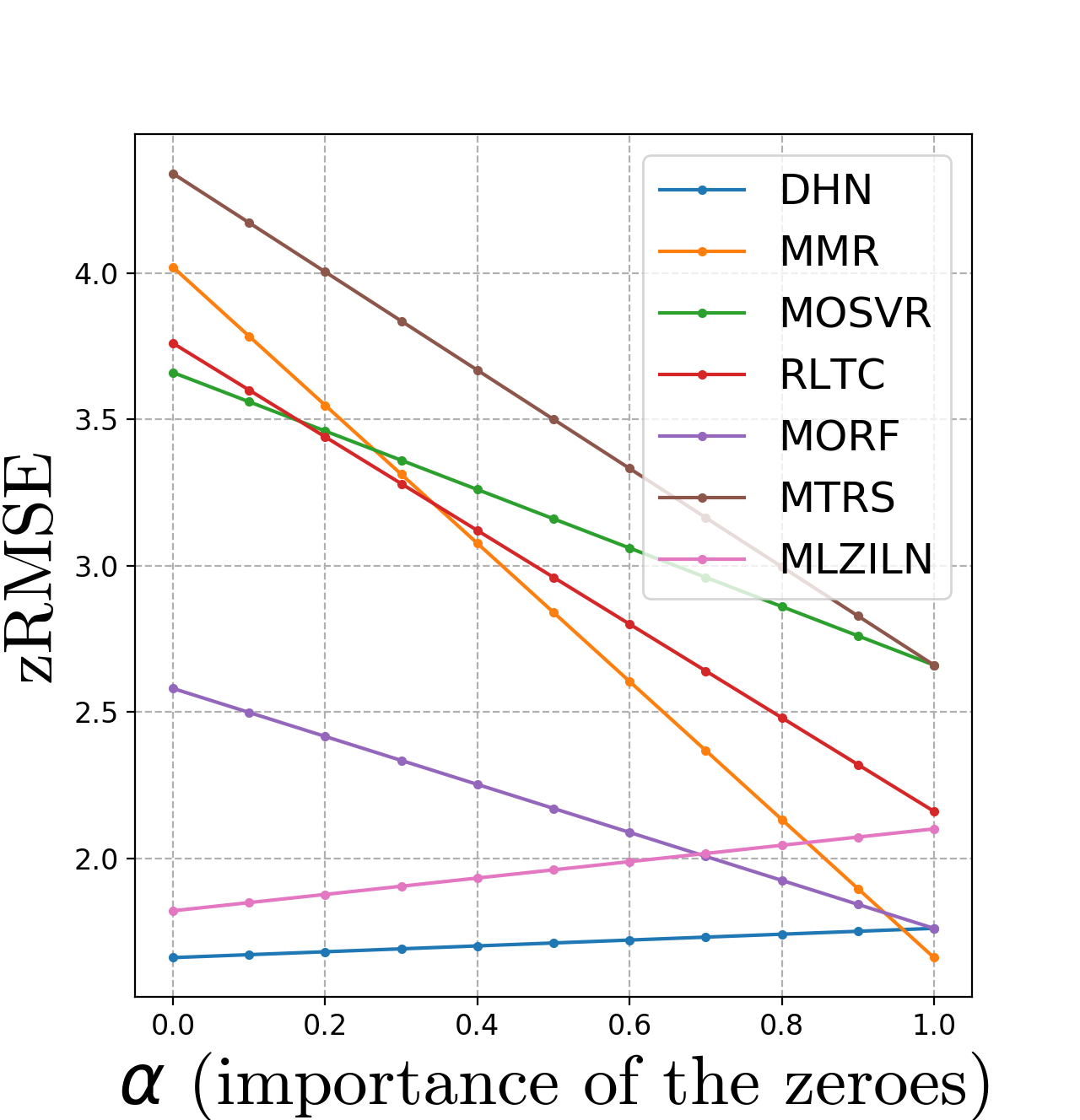}
    \end{minipage}%
    \begin{minipage}{.5\linewidth}
      \centering \includegraphics[width=1.09\linewidth]{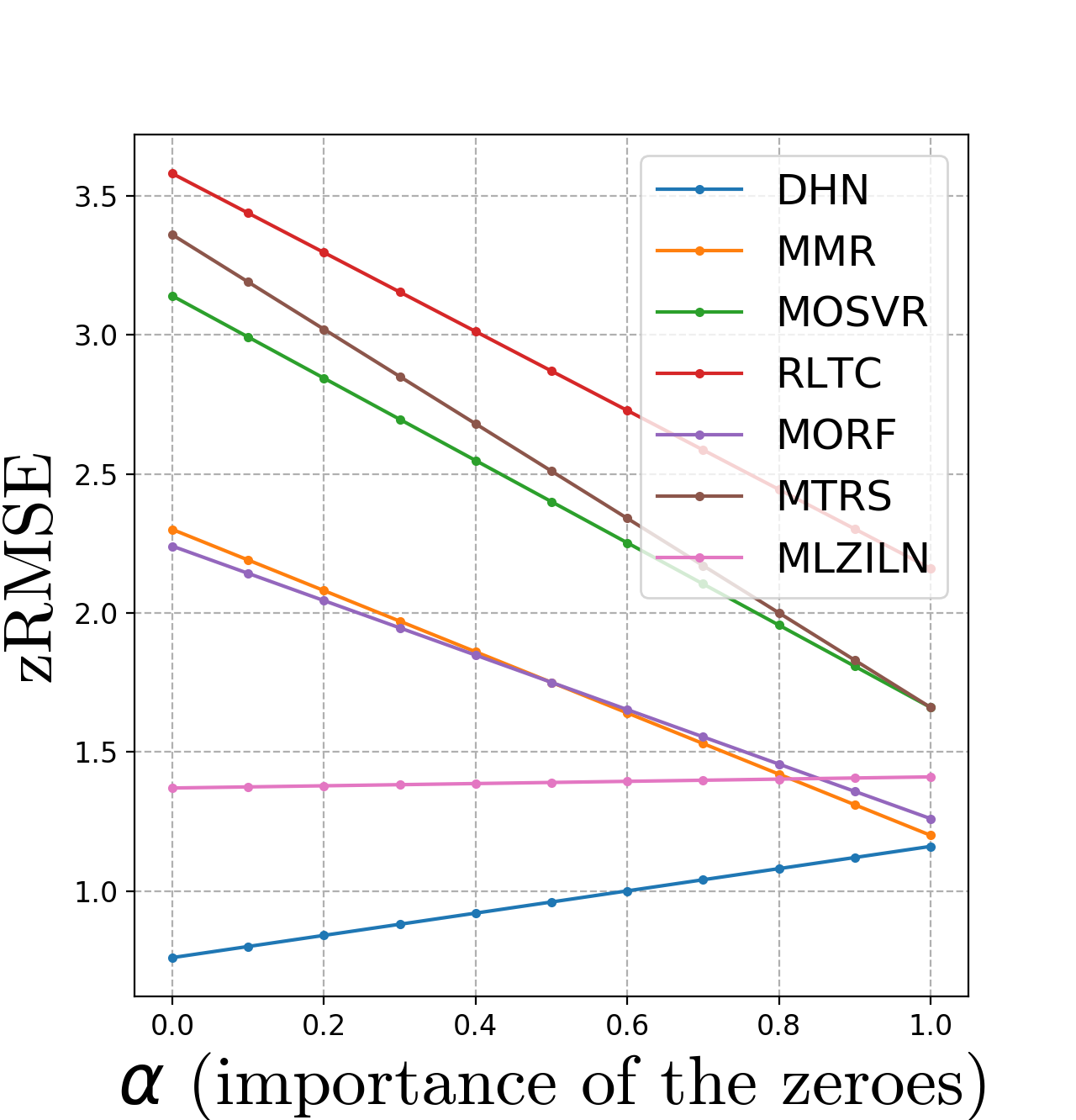}
    \end{minipage}
    \caption{zRMSE of the methods with different $\alpha$ ratios on the SBTS (\textbf{left}) and eBird (\textbf{right}) datasets.
      \label{fig:alpha_zRMSE}}
\end{figure}

\subsection{Implementations and Results}

\begin{small}
\begin{table}
\begin{minipage}[t]{.485\textwidth}
\scalebox{0.85}{
\begin{tabular}{@{}c@{\hspace{0.45em}}c@{\hspace{0.45em}}c@{\hspace{0.45em}}c@{\hspace{0.45em}}c@{\hspace{0.45em}}c@{\hspace{0.45em}}c@{\hspace{0.45em}}|c@{}}
\hline
Metrics & MLZILN & MTRS & MORF & RLTC & MOSVR & MMR & DHN\\
\hline
ACC & 0.52 & 0.31 & 0.45 & 0.40 & 0.32 & 0.47 & \textbf{0.65}\\
zRMSE & 1.96 & 3.50 & 2.17 & 2.96 & 3.16 & 2.84 & \textbf{1.71}\\
Time (min) & 218 & 76 & 87 & 96 & 120 & \textbf{55} & 57\\
\hline
\vspace{0cm}
\end{tabular}
}
\end{minipage}

\begin{minipage}[t]{.485\textwidth}
\scalebox{0.85}{
\begin{tabular}{@{}c@{\hspace{0.45em}}c@{\hspace{0.45em}}c@{\hspace{0.45em}}c@{\hspace{0.45em}}c@{\hspace{0.45em}}c@{\hspace{0.45em}}c@{\hspace{0.45em}}|c@{}}
\hline
Metrics & { }MLZIP{ } & MTRS & MORF & RLTC & MOSVR & MMR & DHN\\
\hline
ACC & 0.50 & 0.32 & 0.41 & 0.19 & 0.39 & 0.27 & \textbf{0.59}\\
zRMSE & 1.39 & 2.51 & 1.75 & 2.87 & 2.40 & 1.75 & \textbf{0.96}\\
Time (min) & 186 & 56 & 64 & 85 & 99 & 53 & \textbf{45}\\
\hline
\end{tabular}
}
\end{minipage}
\caption{Performance comparison between selected models on the SBTS (\textbf{top}) and eBird (\textbf{bottom}) datasets. The best scores are in bold. Except for the MLZILN/MLZIP model, each score is the average after 3 runs. $\alpha$ of zRMSE is set to be 0.5.}
\label{tab:f1s}
\end{table}
\end{small}

Our experiments were carried out on a computer with a 4.2 GHz quad-core Intel i7 CPU, 16 GB RAM and an NVIDIA Quadro P4000 GPU card. We use grid search to find the best hyperparameters for all models, e.g. learning rates, learning rate decay ratio, and tree depth. The encoders of DHN are parametrized by 3-layer fully connected neural networks with latent dimensionalities 512 and 256, and the MLPs of MVP and MLND are 2-layer fully connected neural networks with latent dimensionalities 256. The activation functions in neural networks are set to be ReLU. Neural network models are trained with 100 epochs and batches size 128 for the eBird dataset and 256 for the SBTS dataset. We randomly split the two datasets into three parts for training ($70\%$), validating ($15\%$) and testing ($15\%$), respectively.

The results are shown in Table~\ref{tab:f1s}. As expected, MLZILN/MLZIP and DHN outperform other multi-target regression models in terms of zRMSE because the former models capture zero-inflation of the datasets. DHN has $12.8\%$ and $30.9\%$ lower errors than MLZILN and MLZIP, respectively. On the other hand, DHN has the best ACC on both datasets, which shows the benefit of employing and sharing the same covariance matrix to capture multi-entities correlations. In terms of model training time, MLZILN and MLZIP are far worse than other machine learning models, while DHN and MMR are the best. We also compare zRMSE of the methods under different $\alpha$ ratios in Figure~\ref{fig:alpha_zRMSE}. We can see that zero-inflated models tends to perform better for positive parts, while other nonzero-inflated models tend to underestimate the positive parts. 

\subsection{Ablation Studies}\label{sec:ablation}

To learn the contributions of different components in DHN, we perform simple ablation studies. We modify the DHN architecture and rerun the experiments \blue{in} the following cases: 
\begin{itemize}
\item[1.] We remove the encoder so that both the MVP and MLND directly use the raw features as input. \item[2.] We remove both the encoder and the MVP so that only the MLND is used.
\item[3.] \blue{We do not penalize the difference between the covariance matrices of MVP and MLND.}
\end{itemize}
 
The results are presented in Table~\ref{tab:ablation}. From the results of ablation case studies, we can observe that: (1) employing an encoder to learn latent features helps; (2) the model's performance drops significantly if we do not capture zero inflation of data; and (3) \blue{penalizing the difference between the covariance matrices of} MVP and MLND helps to capture the ACC among \blue{multiple entities} and boost the model's performance.
\begin{small}
\begin{table}
\centering
\scalebox{0.9}{
\begin{tabular}{@{}c@{\hspace{0.45em}}c@{\hspace{0.45em}}c@{\hspace{0.45em}}c@{\hspace{0.45em}}|c@{\hspace{0.45em}}c@{\hspace{0.45em}}c@{\hspace{0.45em}}c@{\hspace{0.45em}}}
\hline
 &  & SBTS &  &  & eBird & \\
\hline
Metrics & Case 1 & Case 2 & Case 3 & Case 1 & Case 2 & Case 3\\
\hline
ACC & 0.65 & 0.50 & 0.45 & 0.59 & 0.47 & 0.43\\
zRMSE & 1.91 & 2.60 & 1.93 & 1.52 & 1.95 & 1.65\\
Time (min) & 55 & 56 & 57 & 44 & 45 & 45\\
\hline
\end{tabular}
}
\caption{Ablation studies. The three cases refer to the ways of modifying our models, which is defined in section~\ref{sec:ablation}. Each score is the average after 3 runs. $\alpha$ of zRMSE is set to be 0.5.}
\label{tab:ablation}
\end{table}
\end{small}

\section{Conclusion}

  To understand the distribution of species across landscapes over time is a key problem in computational sustainability,
  which gives rise to challenging  large-scale  prediction  problems  since hundreds of species have to be simultaneously modeled and the survey data are usually inflated with zeros due to the absence of species for a large number of sites.  We refer to this problem of jointly estimating counts or abundance for multiple entities as zero-inflated multi-target regression.
  
  In this paper,  we have proposed a novel deep model for  zero-inflated multi-target  regression, which is called \blue{the} deep hurdle networks (DHNs).
  \blue{The} DHN simultaneously models zero-inflated data and the correlation among the multiple response variables: 
  we first model the joint distribution of multiple response variables as a multivariate probit model and then couple the positive outcomes with a multivariate log-normal distribution.  
    \blue{A link between both distributions is established by penalizing the difference between their covariance matrices.}
  We then cast the whole model as an end-to-end learning framework  and  provide an efficient learning algorithm for our model that can be fully implemented on GPUs. We show that our model outperforms the existing state-of-the-art baselines on two challenging real-world species  distribution  datasets  concerning  bird  and fish populations. We also performed ablation studies of our models to learn the contributions of different components in the model.  
  
  
  Another related challenging problem in computational sustainability is to forecast how species distribution might be impacted due to the long-term effects of global climate change. In order to tackle the forecast challenge, our future works will consider using recurrent neural networks to improve our model such that it would be able to handle time series data better. On the other hand, we could also borrow ideas of advanced techniques for the multi-label prediction problem to further improve the performance of MVP that is used in our model, e.g. using latent embedding learning to match features and labels in the latent space. 

\section*{Acknowledgments}
This work was partially supported by National Science Foundation OIA-1936950 and CCF-1522054, and the work of Jae Hee Lee was supported by European Research Council Starting Grant 637277. We 
thank the Cornell Lab of Ornithology and Gulf of Maine Research Institute for providing data, resources and advice. Lastly, we thank Richard Bernstein for proofreading the paper.


\bibliographystyle{named}
\bibliography{references}

\end{document}